\documentclass{article}

\usepackage{microtype}
\usepackage{graphicx}
\usepackage{subcaption}
\usepackage{booktabs} 

\usepackage[accepted]{icml2026}
\usepackage{hyperref}

\usepackage{amsmath}
\usepackage{amssymb}
\usepackage{mathtools}
\usepackage{amsthm}

\usepackage[capitalize,noabbrev]{cleveref}

\theoremstyle{plain}

\theoremstyle{definition}

\theoremstyle{remark}

\usepackage[textsize=tiny]{todonotes}

\icmltitlerunning{Certainty-Guided Reasoning in Large
Language Models: A Dynamic Thinking
Budget Approach}

\begin{document}

\twocolumn[
  \icmltitle{Certainty-Guided Reasoning in Large Language Models: A Dynamic Thinking Budget Approach}



  \icmlsetsymbol{equal}{*}

\begin{icmlauthorlist}
\icmlauthor{Jo\~ao Paulo Nogueira}{ipp,nbl}
\icmlauthor{Wentao Sun}{ep,nbl}
\icmlauthor{Alonso Silva}{nbl}
\icmlauthor{Laith Zumot}{nokia}
\end{icmlauthorlist}

\icmlaffiliation{ipp}{Institut Polytechnique de Paris, Paris, France}
\icmlaffiliation{ep}{\'Ecole Polytechnique, Paris, France}
\icmlaffiliation{nokia}{Nokia, France}
\icmlaffiliation{nbl}{Nokia Bell Labs, France}

\icmlcorrespondingauthor{Jo\~ao Paulo Nogueira}{joaopaulo.fontoura@ip-paris.fr}
\icmlcorrespondingauthor{Alonso Silva}{alonso.silva@nokia-bell-labs.com}

  \icmlkeywords{Machine Learning, NLP, Reasoning, LLM}

  \vskip 0.3in
]



\printAffiliationsAndNotice{}  

\begin{abstract}
Large reasoning language models are typically run with fixed inference budgets, which can waste computation or terminate reasoning prematurely. We introduce Certainty-Guided Reasoning (CGR), a model-agnostic adaptive inference procedure that periodically probes whether the current reasoning supports a confident final answer and terminates early once a target certainty threshold is reached, otherwise continuing until the end-of-thinking token or the budget limit. Certainty is estimated from the model’s predicted probabilities over the  answer tokens, yielding a lightweight stopping criterion. On AIME2025, CGR preserves baseline accuracy while reducing token usage, providing a tunable certainty–efficiency trade-off that can eliminate millions of tokens in aggregate. Across 64 random seeds, CGR exhibits consistent behavior. We also introduce a Grade metric that penalizes incorrect answers and permits abstention, capturing risk-sensitive performance. Results show that CGR improves Grade by abstaining when certainty remains low.

\end{abstract}

\section{Introduction}
Large reasoning language models (LRLMs) can solve problems that require multi step
deliberation, yet they are commonly run with a fixed inference budget that assigns the same
amount of reasoning to every query. This choice creates a mismatch between allocated compute and per instance deliberation requirements, which wastes tokens on instances that require little reasoning and truncates instances that require extended inference. The mismatch also matters empirically since test
time compute does not always improve performance and can exhibit adverse scaling behavior \citep{gema2025inverse}.

We introduce \textbf{Certainty Guided Reasoning (CGR)}, a \textbf{model-agnostic} inference
procedure that uses a model's own confidence signal to control how long it reasons. CGR probes
the current reasoning state at fixed intervals by eliciting a candidate final answer and then
computes a certainty score from the model's predicted probabilities of the answer tokens. When
certainty exceeds a threshold, CGR stops and returns the answer. When certainty remains below
the threshold, CGR continues reasoning until it reaches the maximum budget. This setup can be
viewed as a generator and critic interaction in which generation produces a reasoning trace and
probing evaluates whether the trace is sufficient to stop \citep{goodfellow2014gan}. In our
experiments, the same model implements both roles, although the framework also permits a
separate probing model.

CGR builds on the premise that internal uncertainty is informative for control. If a model can
reliably detect low certainty cases, it can avoid committing to unreliable outputs while spending
additional compute only when it is needed. This premise aligns with evidence that language
models often contain usable signals about what they know \citep{kadavath2022know}. CGR
turns that signal into an explicit stopping rule without modifying model weights.

We evaluate CGR on AIME2025 across multiple open weight LRLMs and we report results over
64 random seeds. CGR preserves baseline accuracy while reducing token usage, and it yields
more consistent outcomes across seeds than fixed budget baselines. We also introduce a Grade
metric that penalizes incorrect answers and permits abstention, which lets us evaluate
risk sensitive performance under different penalty regimes.

Our main contributions are
\begin{itemize}
    \item We propose an active certainty probing mechanism that periodically simulates a final
    answer and converts answer token probabilities into a stopping criterion.
    \item We formulate CGR as a model-agnostic inference method that requires no specialized
    training and no architectural modification.
    \item We demonstrate that CGR generalizes across model scales and preserves baseline
    accuracy while reducing token usage on AIME2025, with improved stability over 64 seeds.
    \item We introduce a penalty based Grade metric and show that CGR improves Grade by
    abstaining when certainty remains low.
\end{itemize}

\section{Related Work}
Our work builds on three lines of research that study how to allocate test time computation in
language models and how to use model uncertainty as a control signal.

\paragraph{Dynamic test time compute and early exit.}
Early exit methods allocate less computation to inputs that do not require full depth processing.
DeeBERT introduces confidence gated exits at intermediate Transformer layers to accelerate
inference while maintaining accuracy \citep{xin2020deebert}. Recent work extends the same
principle to reasoning models where computation unfolds across generated reasoning tokens
rather than layers. DEER monitors generation with either explicit reasoning markers or entropy
based signals and terminates reasoning when confidence is high \citep{yang2025deer}. CGR
differs in its stopping mechanism. Instead of relying on passive markers or token level entropy,
it periodically performs an explicit probe that simulates a candidate final answer and evaluates
certainty from the answer token probabilities.

\paragraph{Test time scaling and controlling reasoning length.}
A complementary direction increases test time compute to improve reasoning performance.
s1 demonstrates that additional test time computation can be a strong driver of reasoning
quality and introduces budget forcing as a control mechanism \citep{muennighoff2025s1}.
At the same time, increasing reasoning length does not guarantee monotonic gains and can
lead to overthinking effects \citep{gema2025inverse}. CGR targets this tension by using
certainty to allocate compute bidirectionally. It can extend reasoning when confidence remains
low while also terminating when a stable high certainty answer emerges. Budget guidance
offers another approach that uses an auxiliary predictor to steer generation toward a target
budget \citep{li2025budgetguidance}. CGR differs by using an intrinsic signal derived from the
model during inference and it does not require an external predictor.

\paragraph{Uncertainty estimation and confidence guided reasoning.}
A substantial body of work studies how to quantify uncertainty in large language models and
how to use it for reliable decision making. \citet{shorinwa2025survey} survey uncertainty estimation
methods and highlight open challenges in treating token probabilities as calibrated confidence.
\citet{kadavath2022know} provide evidence that language models often
exhibit informative signals about what they know. Confidence has
also been used as a training signal. RENT improves reasoning by reinforcing trajectories that
reduce entropy without external supervision \citep{prabhudesai2025rent}. CGR uses confidence
differently. It treats certainty as an inference time control signal that determines when to stop
reasoning, rather than as a reward used during training.

\section{Method}
We study inference for a reasoning language model that produces a sequence of reasoning tokens
followed by a final answer. In many LRLMs this process is delimited by a special token that
signals the end of reasoning. We treat \texttt{</think>} as the end of thinking token.

Let $q$ denote the input query and let $B$ denote a maximum thinking token budget. The model
generates tokens autoregressively to form a partial output $o_{1:t}$. CGR adds an explicit
stopping rule that can terminate generation before the budget is exhausted when the model can
already produce a sufficiently certain final answer.

\subsection{Certainty estimation}
A cornerstone of CGR is quantifying the model's certainty in a proposed final answer. This
signal is derived from next token prediction in an autoregressive language model. At each
generation step $i$, the model $M$ produces logits
$\mathbf{z}_i = M(q, o, t_{<i})$ given the query $q$, the current trace $o$, and the previously
generated tokens $t_{<i}$. Applying a softmax yields a distribution over the vocabulary, and
the probability of generating a specific next token $t_i$ is
\begin{equation}
p(t_i \mid q, o, t_{<i}) = \mathrm{softmax}(\mathbf{z}_i)[t_i] .
\end{equation}

We use these token probabilities to define the certainty of a multi token answer produced by a
probe. During probing, $o$ denotes the current trace augmented with a fixed answer prefix. The
probe decodes an answer greedily under this prefix. At each answer step $i$, it selects the
most likely next token,
\begin{equation}
t_i^\star = \arg\max_{v \in \mathcal{V}} p(v \mid q, o, t_{<i}^\star) .
\end{equation}
Let $a^\star = (t_1^\star, \ldots, t_n^\star)$ denote the resulting answer token sequence,
excluding fixed formatting delimiters. We define the certainty of $a^\star$ as the minimum
probability assigned to any selected answer token,
\begin{equation}
c(a^\star) = \min_{i \in \{1,\ldots,n\}} p(t_i^\star \mid q, o, t_{<i}^\star) .
\label{eq:certainty}
\end{equation}

This minimum aggregation is intentionally conservative. It requires every answer token to be
predicted with high confidence, which is particularly important for short numeric answers where
a single uncertain digit invalidates the entire output. Alternative aggregations such as the
mean or product can be dominated by a subset of high confidence tokens and can therefore mask
localized uncertainty.

\subsection{Certainty Guided Reasoning}
CGR alternates between generating thinking tokens and periodically checking whether a high
certainty final answer is already available. The procedure uses two hyperparameters. The probe
interval $\Delta$ determines how often CGR evaluates certainty. The threshold $\theta$ sets the
minimum acceptable certainty for early termination.

At each step, the model generates the next token. If it emits \texttt{</think>}, CGR stops
thinking and proceeds to decode the final answer. Otherwise, every $\Delta$ thinking tokens CGR
runs a certainty probe, computes $c(a^\star)$ from Equation~\ref{eq:certainty}, and stops early
when $c(a^\star) \ge \theta$. If neither condition is met, generation continues until the
budget $B$ is reached.

\paragraph{Answer prefix for probing.}
CGR evaluates certainty by forcing the model into an explicit answer mode. At each probe time
we append a fixed answer prefix to the current trace,
\texttt{Final Answer: \textbackslash boxed\{} ,
and then greedily decode the ensuing answer tokens. This prefix standardizes the probe context
across time steps and ensures that the certainty score is computed on the same answer format
that would be produced if CGR terminated at that point.

\begin{algorithm}[tbh]
  \caption{Certainty Guided Reasoning}
  \label{alg:cgr}
  \begin{algorithmic}
    \STATE {\bfseries Input:} query $q$, model $M$, budget $B$, threshold $\theta$, interval $\Delta$
    \STATE {\bfseries Probe prefix:} \texttt{Final Answer: \textbackslash boxed\{}
    \STATE $o \leftarrow \emptyset$
    \FOR{$t=1$ {\bfseries to} $B$}
      \STATE Sample next token $x \sim M(\cdot \mid q, o)$
      \STATE $o \leftarrow o \Vert x$
      \IF{$x$ equals \texttt{</think>}}
        \STATE {\bfseries break}
      \ENDIF
      \IF{$t \bmod \Delta = 0$}
        \STATE $c \leftarrow \textsc{CertaintyProbe}(M, q, o)$
        \IF{$c \ge \theta$}
          \STATE {\bfseries break}
        \ENDIF
      \ENDIF
    \ENDFOR
    \STATE {\bfseries Output:} $\textsc{DecodeAnswer}(M, q, o)$
  \end{algorithmic}
\end{algorithm}

\paragraph{CertaintyProbe and DecodeAnswer.}
Both functions append the probe prefix to the current trace and greedily decode answer tokens.
\textsc{CertaintyProbe} records the probability of each selected answer content token and
returns the minimum over those probabilities. The minimum excludes fixed formatting delimiters.
\textsc{DecodeAnswer} returns the decoded answer string.

\subsection{Optional probing model}
CGR does not require the same model to both generate thinking tokens and evaluate certainty.
In our experiments the same model performs both roles for simplicity and to avoid introducing
additional components. The same procedure permits a distinct probing model $M_{\text{probe}}$
that consumes the current trace and produces the answer token probabilities used in
Equation~\ref{eq:certainty}. This variant can be useful when a smaller model is used for
probing or when the generation model does not expose token probabilities directly.

\section{Experimental setup}
\subsection{Datasets}
We evaluate on AIME2025, which contains 30 competition mathematics
problems from the American Invitational Mathematics Examination. \citep{maaAIME}
Each problem requires multi-step reasoning, and the ground truth answer is a single integer in
$[0,999]$. 
For automatic scoring, we require the model to output the final answer in a boxed format. We
treat an output as valid only if the extracted answer is a numeric integer that exactly matches
the ground truth. 

\subsection{Models}
To reduce the risk of dataset contamination, we use open-weight reasoning models whose
training cutoffs predate the release of AIME2025. We evaluate DeepSeek-R1-Distill-Qwen-14B
and DeepSeek-R1-Distill-Llama-70B, both distilled from DeepSeek-R1, and Phi-4-reasoning-plus.
\citep{deepseekR1,abdin2025phi4}

\subsection{Inference protocol and baselines}
All experiments use zero-shot chain-of-thought prompting in the style of \citet{wei2022cot}.
We run decoding at temperature $0.6$ and provide the instruction
\texttt{Please reason step by step, and put your final answer within \textbackslash boxed\{\}}.
For the fixed-budget baseline, the model generates thinking tokens until it emits
\texttt{</think>} or reaches a specified thinking budget. We then append the answer prefix
\texttt{Final Answer: \textbackslash boxed\{} to force answer generation.

To characterize performance as a function of compute, we sweep fixed thinking budgets from
$1{,}000$ to $32{,}000$ tokens in increments of $1{,}000$. CGR uses the same prompting and
answer extraction procedure, and applies probing every $1{,}000$ thinking tokens unless noted
otherwise.

\subsection{Reproducible evaluation via post hoc simulation}
A direct real-time implementation of probing introduces non-determinism in long generations.
Even with identical prompts and seeds, small floating point differences can accumulate and
produce different reasoning traces.
To isolate the effect of certainty thresholds, we adopt a post hoc evaluation protocol.

First, for each question and random seed, we generate and save a full $32{,}000$ token thinking
trace. Second, we simulate early stopping by cutting the saved trace at every $1{,}000$ token
boundary and running the probe on the truncated trace.
This ensures that all threshold comparisons share the same underlying reasoning, and that
performance differences are attributable to the stopping rule rather than to stochastic trace
variation.

\subsection{Probe decoding details}
At each probe point, we append the fixed answer prefix and greedily decode a short answer
sequence. In our implementation, probe answer decoding is capped at $L=4$ tokens, which is sufficient for AIME answers. For each model tokenizer, we verified all integers 0–999 fit within L tokens under our answer format. The certainty score is computed over answer content tokens and
excludes fixed formatting delimiters.

\section{Metrics}
We report three metrics that capture correctness, risk sensitive utility, and computational
efficiency.

\subsection{Accuracy}
Accuracy is the fraction of questions answered correctly. A response is counted as correct if
and only if the extracted final answer is a numeric integer that exactly matches the ground
truth solution.

\subsection{Grade}
To evaluate risk sensitive behavior, we define a Grade metric inspired by exam scoring rules
that penalize incorrect answers and allow abstention. For each query, the score is
\begin{equation}
\mathrm{Grade}(q) =
\begin{cases}
+1 & \text{if the answer is correct} \\
0  & \text{if the system abstains} \\
-p & \text{if the answer is incorrect}
\end{cases}
\end{equation}
where $p \ge 0$ is a penalty coefficient. We evaluate $p \in \{0, 0.25, 0.5, 1.0\}$ to represent
no penalty, mild penalty, moderate penalty, and strong penalty regimes.

We implement abstention using the same certainty signal as CGR. At a probe point or at the end
of the budget, if the predicted answer certainty $c(a^\star)$ is below the threshold $\theta$,
the system abstains. Otherwise it outputs the decoded answer. Dataset level Grade is computed
as
\begin{equation}
\mathrm{Grade}(\mathcal{D}) = N_{\mathrm{correct}} - p \, N_{\mathrm{incorrect}} .
\end{equation}
The baseline Grade corresponds to answering all questions without abstention.

\subsection{Token savings}
We quantify computational efficiency by token savings relative to a fixed budget baseline. For
a given query, let $T_{\mathrm{base}}$ denote the number of thinking tokens generated by the
baseline run under its budget and let $T_{\mathrm{cgr}}$ denote the number of thinking tokens
generated before CGR terminates. Token savings are defined as
\begin{equation}
\mathrm{Saved}(q) = T_{\mathrm{base}} - T_{\mathrm{cgr}} .
\end{equation}
We report aggregate savings across a dataset as $\sum_{q \in \mathcal{D}} \mathrm{Saved}(q)$ and
we also report averages per seed and per question.

\section{Results}
We evaluate CGR on AIME2025 across DeepSeek 14B, Phi 4, and DeepSeek 70B. Unless stated
otherwise, results aggregate 64 random seeds and probing is performed every 1{,}000 thinking
tokens.

\subsection{Baseline performance}
Table~\ref{tab:baseline} reports fixed budget baseline accuracy at 32{,}000 thinking tokens.
These baselines establish the attainable reasoning performance for each model under the same
budget and prompt format.

\begin{table}[th]
\caption{Baseline performance on AIME2025 at 32{,}000 thinking tokens.}
\label{tab:baseline}
\centering
\begin{tabular}{lcc}
\toprule
Model & Accuracy (out of 30) & Accuracy (\%) \\
\midrule
DeepSeek 14B & 14.31 & 47.70 \\
DeepSeek 70B & 15.43 & 51.45 \\
Phi 4 & 22.17 & 73.90 \\
\bottomrule
\end{tabular}
\end{table}

\subsection{Accuracy under certainty thresholds}
We sweep certainty thresholds $\theta \in \{0.96, 0.97, 0.98, 0.99\}$ and compare against the
fixed budget baseline.

\paragraph{DeepSeek 14B.}
Table~\ref{tab:ds14b_thresholds} shows that accuracy remains close to baseline across
thresholds, with the strongest setting $\theta = 0.99$ within 1.1\% relative of baseline. The
distribution of per seed differences concentrates near zero at higher thresholds, indicating
reduced variance and fewer large deviations. This pattern is consistent with CGR abstaining on
low certainty instances rather than producing additional incorrect answers.

\begin{table}[th]
\caption{CGR on DeepSeek 14B across certainty thresholds on AIME2025 over 64 seeds.}
\label{tab:ds14b_thresholds}
\centering
\begin{tabular}{lccccc}
\toprule
$\theta$ & Acc. & Rel. Acc. & Acc. Diff. & Equal & Better \\
\midrule
0.96 & 13.41 & 44.69 & -0.92 & 18 & 7 \\
0.97 & 13.58 & 45.26 & -0.75 & 20 & 9 \\
0.98 & 13.77 & 45.89 & -0.56 & 23 & 9 \\
0.99 & 13.98 & 46.61 & -0.34 & 31 & 8 \\
\bottomrule
\end{tabular}
\end{table}

\paragraph{Phi 4.}
Table~\ref{tab:phi4_thresholds} shows that Phi 4 accuracy is effectively invariant across
thresholds. The relative difference remains under 1\% and the majority of seeds match baseline.
This suggests that Phi 4 tends to produce strongly polarized certainty values, which reduces
the impact of threshold tuning.

\begin{table}[th]
\caption{CGR on Phi 4 across certainty thresholds on AIME2025 over 64 seeds.}
\label{tab:phi4_thresholds}
\centering
\begin{tabular}{lccccc}
\toprule
$\theta$ & Acc. & Rel. Acc. & Acc. Diff. & Equal & Better \\
\midrule
0.96 & 21.89 & 72.97 & -0.29 & 45 & 0 \\
0.97 & 21.89 & 72.97 & -0.29 & 45 & 0 \\
0.98 & 21.89 & 72.97 & -0.29 & 45 & 0 \\
0.99 & 21.89 & 72.97 & -0.29 & 45 & 0 \\
\bottomrule
\end{tabular}
\end{table}

\paragraph{DeepSeek 70B.}
For DeepSeek 70B, CGR closely mirrors baseline accuracy with minimal divergence. Runs that
deviate include both slight improvements and slight deteriorations, and the distance from
baseline remains small, indicating stable behavior at scale.

\subsection{Answer progression over thinking budget}
To understand when correct answers emerge, we analyze cumulative correct and incorrect
predictions as a function of available thinking tokens. For DeepSeek 14B, correct answers accumulate steadily up to roughly 10{,}000 tokens and then flatten, while the ratio of incorrect to correct answers decreases after roughly 15{,}000 tokens and gains become marginal beyond 25{,}000 tokens. 
DeepSeek 70B exhibits a similar trend, hitting a plateau at 20{,}000 tokens. 
This supports periodic probing at 1{,}000 token intervals since it captures the
dominant transitions while keeping overhead small. For Phi 4, most correct answers appear before 10{,}000 tokens and performance plateaus thereafter, indicating earlier convergence.

\begin{figure}[t]
  \centering
  \includegraphics[width=\linewidth]{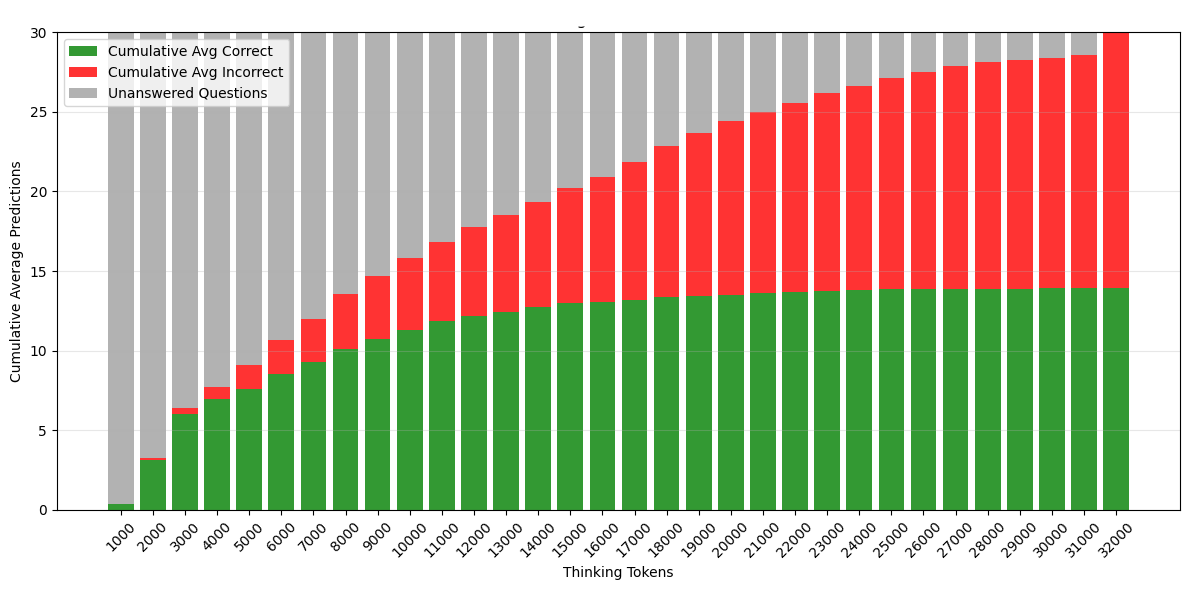}
    \caption{DeepSeek 14B cumulative predictions over the thinking budget.}
  \label{fig:ds14b_answers}
\end{figure}

\begin{figure}[h]
  \centering
  \includegraphics[width=\linewidth,trim={0 0 0 22pt},  clip]{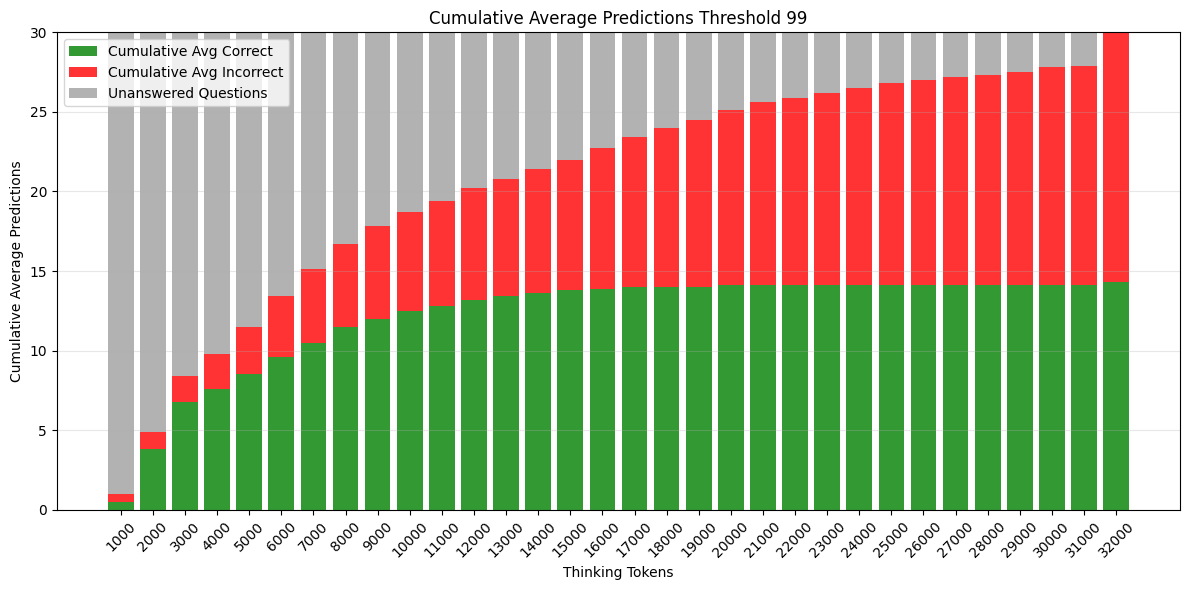}
  \caption{Deepseek 70b cumulative predictions over the thinking budget.}
  \label{fig:deep70_answers}
\end{figure}

\begin{figure}[htb]
  \centering
  \includegraphics[width=\linewidth]{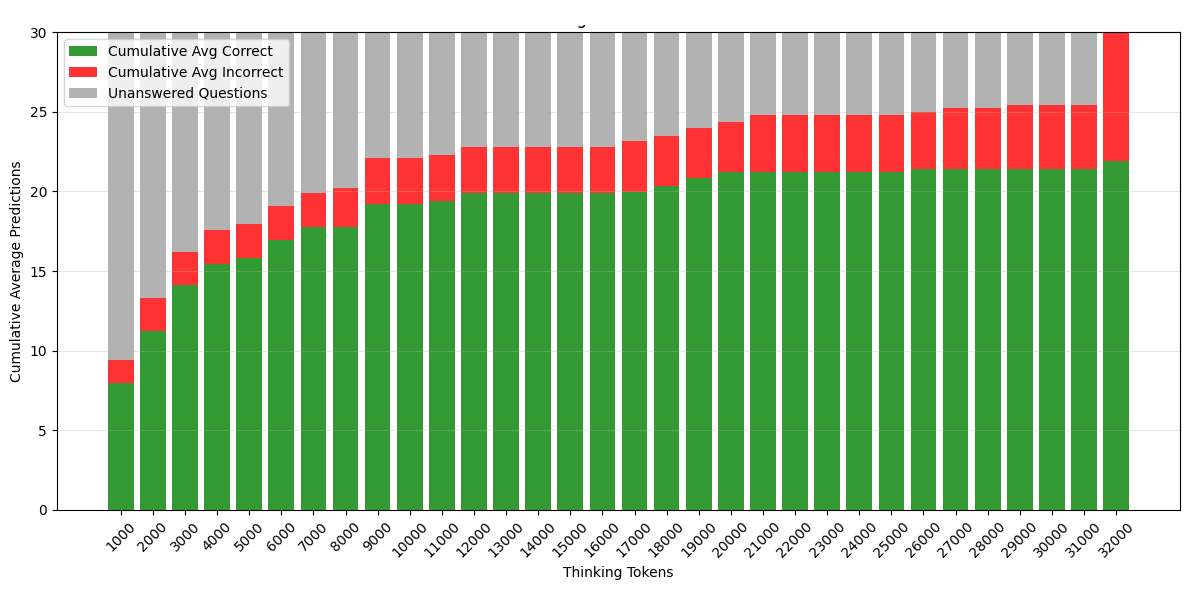}
  \caption{Phi 4 cumulative predictions over the thinking budget.}
  \label{fig:phi4_answers}
\end{figure}

The apparent jump near 32{,}000 tokens in Figures~\ref{fig:ds14b_answers}, \ref{fig:deep70_answers}, and ~\ref{fig:phi4_answers} reflects residual low certainty cases that did not satisfy the threshold before the budget limit. These late completions were retained in post-processing so that CGR can be compared directly against full-budget baselines at 32{,}000 tokens. This pattern is consistent with adaptive compute allocation, since simpler instances typically reach stable certainty within the first few thousand tokens and terminate early, while more complex instances
continue reasoning until later intervals.

\begin{figure}[h]
  \centering
  \includegraphics[width=\linewidth,trim={0 0 0 22pt},  clip]{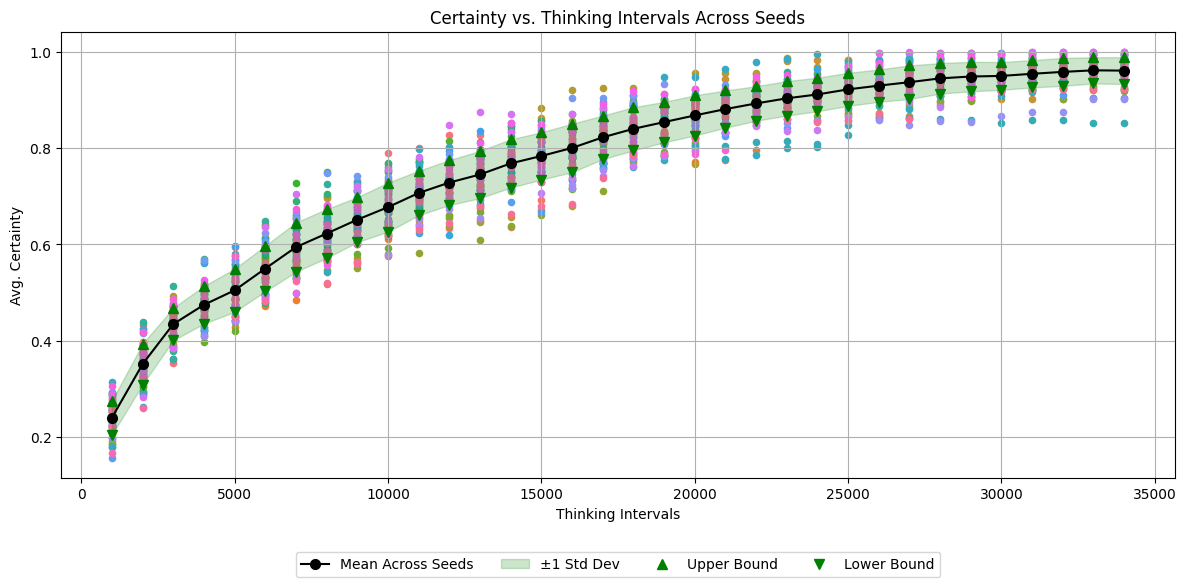}
  \caption{Average certainty at each step.}
  \label{fig:avg_cert}
\end{figure}

As illustrated in Figure \ref{fig:avg_cert}, this performance plateau is mirrored by the evolution of the model’s internal certainty signal. Across 64 random seeds, the mean certainty score exhibits a monotonic increase as the thinking budget expands. While early reasoning intervals are characterized by low certainty (typically between 0.2 and 0.4) and high variance across seeds, the signal eventually saturates as the reasoning trace lengthens. This suggests that as the context becomes increasingly specified through extended deliberation, the probability distribution over the final answer tokens becomes highly polarized. In this regime, additional computation primarily serves to reinforce the existing consensus within the trace rather than surfacing new correct answers, which would explain why the cumulative accuracy in Figures~\ref{fig:ds14b_answers}, \ref{fig:deep70_answers}, and \ref{fig:phi4_answers} flatten even as certainty continues to rise toward 1.0.

\subsection{Grade under penalty based scoring}
We evaluate risk sensitive performance using Grade, which penalizes incorrect answers and
permits abstention. Grade exposes behavior that accuracy alone cannot capture by rewarding caution when penalties are significant.

\paragraph{DeepSeek 14B.}
Table~\ref{tab:ds14b_grade} reports Grade at 31{,}000 tokens for $\theta = 0.99$ and the
token step at which Grade is maximized. Under penalties $p \in \{0.5, 1.0\}$, CGR improves
Grade relative to baseline, and the maximizing step occurs substantially earlier than the full
budget.

\begin{table}[ht]
\caption{DeepSeek 14B Grade at 31{,}000 tokens and maximal Grade step for $\theta=0.99$.}
\label{tab:ds14b_grade}
\centering
\resizebox{\columnwidth}{!}{%
\begin{tabular}{lccccc}
\toprule
$p$ & Final Grade & Baseline & Diff. & Max Grade & Step \\
\midrule
0.00 & 13.98 & 14.31 & -0.34 & 13.98 & 31{,}000 \\
0.25 & 10.28 & 10.39 & -0.11 & 11.18 & 15{,}000 \\
0.50 & 6.59  & 6.47  & 0.12  & 9.42  & 12{,}000 \\
1.00 & -0.78 & -1.38 & 0.59  & 6.91  & 11{,}000 \\
\bottomrule
\end{tabular}%
}
\end{table}

\paragraph{Phi 4.}
Table~\ref{tab:phi4_grade} shows that the advantage of abstention becomes more pronounced as
penalties increase. At $p = 1.0$, CGR improves Grade relative to baseline and reaches its
maximum Grade at a shorter budget than 32{,}000.

\begin{table}[t]
\caption{Phi 4 Grade at 31{,}000 tokens and maximal Grade step for $\theta=0.99$.}
\label{tab:phi4_grade}
\centering
\resizebox{\columnwidth}{!}{%
\begin{tabular}{lccccc}
\toprule
$p$ & Final Grade & Baseline & Diff. & Max Grade & Step \\
\midrule
0.00 & 21.39 & 22.17 & -0.78 & 21.39 & 26{,}000 \\
0.25 & 20.38 & 20.21 & 0.16  & 20.50 & 26{,}000 \\
0.50 & 19.36 & 18.26 & 1.10  & 19.62 & 20{,}000 \\
1.00 & 17.33 & 14.34 & 2.98  & 18.03 & 20{,}000 \\
\bottomrule
\end{tabular}%
}
\end{table}


\subsection{Token savings}
CGR reduces computation by exiting early when certainty exceeds the threshold. Table
\ref{tab:token_savings1} reports total thinking tokens saved for DeepSeek 14B relative to a
32{,}000 token baseline. Lower thresholds produce larger savings, while higher thresholds are
more conservative and save fewer tokens. Even at $\theta = 0.99$, CGR saves over two million
tokens in aggregate across 64 seeds.

\begin{table}[th]
\caption{DeepSeek 14B token savings on AIME2025 across thresholds over 64 seeds.}
\label{tab:token_savings1}
\centering
\begin{tabular}{lccc}
\toprule
$\theta$ & Total saved & Avg. per seed & Avg. per question \\
\midrule
0.96 & 3{,}380{,}578 & 52{,}821 & 1{,}760 \\
0.97 & 3{,}081{,}690 & 48{,}151 & 1{,}605 \\
0.98 & 2{,}739{,}761 & 42{,}808 & 1{,}426 \\
0.99 & 2{,}042{,}389 & 31{,}912 & 1{,}063 \\
\bottomrule
\end{tabular}
\end{table}

\begin{figure}[th]
\label{token_savings}
  \centering
  \includegraphics[width=\linewidth]{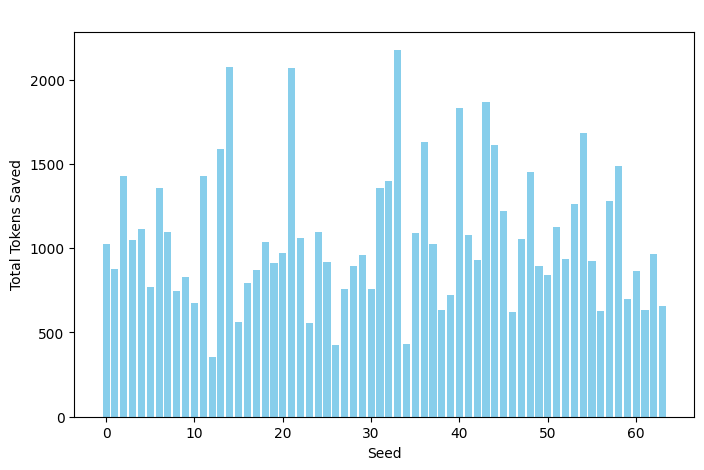}
  \caption{Token savings per seed with CGR token usage.}
  \label{fig:token_savings}
\end{figure}

Figure~\ref{fig:token_savings} breaks token savings down by random seed for $\theta = 0.99$.
Savings vary across seeds, which reflects differences in how quickly the model reaches the
certainty threshold on a given run. Despite this variability, most seeds exhibit substantial
early exits, which explains the multi million token reduction reported in
Table~\ref{tab:token_savings1}.

\section{Ablations}
\subsection{Certainty threshold trade-off}
CGR exposes a controllable trade-off between efficiency and conservatism through the certainty
threshold $\theta$. Lower thresholds trigger earlier exits and therefore increase token savings,
but they can also forgo late emerging correct answers. Higher thresholds reduce the frequency
of early exits, which decreases savings, but they concentrate decisions on cases where the model
assigns consistently high probability to the full answer token sequence. In our experiments,
$\theta = 0.99$ provides the most robust operating point for risk averse use. It remains close to
baseline accuracy while still delivering substantial aggregate token savings.

\subsection{Probing every 1{,}000 tokens}
Certainty probing introduces additional computation because it requires decoding an answer
under the probe prefix and recording token probabilities for the selected answer tokens. If
probing is performed at very high frequency, probe overhead can dominate total inference cost
and can negate the efficiency gains of early exiting.

We therefore probe at a coarse interval that provides visibility into reasoning progress while
keeping overhead small. An interval of 1{,}000 thinking tokens yields at most 32 probe calls
under a 32{,}000 token budget, which bounds probe cost by a small constant factor. This
granularity is also compatible with observed reasoning dynamics in our experiments, where
performance changes occur over thousands of tokens rather than at single token resolution.

\subsection{Budget forcing}
Budget forcing is a test time control mechanism that prevents a model from terminating its
reasoning before reaching a predetermined token budget \citep{muennighoff2025s1}.
 CGR's early exit
improves efficiency by terminating once the model reaches a stable, high certainty answer. We combined these methods into a variant that integrates budget forcing with CGR. In this variant, the
model is compelled to continue thinking until it either reaches the budget limit or produces an
answer whose certainty exceeds the threshold. Following prior work on test time scaling, we
implement budget forcing by suppressing an early \texttt{</think>} and replacing it with a
continuation token such as \texttt{\textbackslash nWait} \citep{muennighoff2025s1}. 

Across the evaluated models, budget forcing provides limited practical benefit. For DeepSeek
14B it largely matches baseline accuracy, which suggests that additional forced computation
often yields diminishing returns. For DeepSeek 70B it can degrade performance, consistent with
reasoning drift induced by unnecessary continuation. In all cases, budget forcing conflicts with
the efficiency objective because it eliminates most of the token savings that CGR achieves via
early exits.

\section{Discussion and Conclusion}
Certainty Guided Reasoning operationalizes an inference time control principle that is simple and model agnostic. It treats internal answer certainty as a signal for reasoning sufficiency and uses periodic probing to decide whether additional thinking tokens are warranted. This converts a fixed budget procedure into an adaptive one, where compute is allocated in response to the model’s own uncertainty rather than a uniform rule.

Across AIME2025, CGR preserves baseline accuracy while reducing the average number of generated tokens, and the threshold parameter provides a direct certainty efficiency trade off. The multi seed evaluation indicates that this behavior is stable, which matters for reproducibility when results depend on sampling noise. The Grade metric further highlights a practical regime that is not captured by accuracy alone, since abstaining on low certainty instances can improve effective performance when incorrect answers carry explicit penalties.

The ablation results clarify the boundary conditions of the approach. Early exit is most beneficial when the model reaches a high certainty answer well before the budget, whereas forcing additional reasoning can erase token savings and can degrade outcomes for larger models whose default stopping behavior is already well calibrated. This suggests that test time controls should be treated as complementary tools rather than universally additive improvements, and that the interaction between stopping and forcing deserves explicit measurement in any deployment setting.

This outcome is consistent with our answer progression analysis. For DeepSeek 14B, the number
of correct answers increases rapidly up to approximately 10{,}000 thinking tokens and then the
slope flattens, while gains beyond roughly 25{,}000 tokens are marginal. At the same time, the
error profile continues to evolve, since the ratio of incorrect to correct answers decreases only
after roughly 15{,}000 tokens. These dynamics clarify the limitation of the budget forcing
premise that more thinking monotonically improves outcomes. Additional compute can recover a
small number of late emerging correct answers, but the remaining unresolved instances are
increasingly dominated by failures, which raises a practical question of whether the marginal
accuracy gains justify the added inference cost.

Several extensions are immediate. The certainty function could be generalized beyond the minimum token probability to incorporate calibration techniques or alternative aggregations that better reflect semantic uncertainty. Probing policies could also be adapted, for example by varying the probe interval as a function of observed certainty dynamics, and a dedicated probing model could reduce overhead when generation models do not expose reliable token probabilities. Finally, while AIME2025 provides a clean mathematical benchmark, a broader evaluation across domains and answer formats would better characterize when certainty based control transfers and when it fails.

Overall, CGR demonstrates that internal confidence is a useful control signal for reasoning length, and it provides a lightweight mechanism for improving resource efficiency while promoting more reliable behavior under uncertainty.

\section*{Impact Statement}
Certainty Guided Reasoning is intended to improve the efficiency and reliability of inference in
reasoning language models by adaptively terminating computation when the model is sufficiently
confident. In large scale deployments, earlier termination can reduce inference cost and energy
use. The same certainty signal can also support selective answering in settings where abstention
is preferable to low confidence outputs, which is relevant when incorrect answers carry explicit
penalties.

The main risk is that token probabilities may not be well calibrated under distribution shift or
prompt changes, which can lead to overly conservative abstention or occasional high confidence
errors. In practice, this can be mitigated by validating thresholds on in domain data, monitoring
abstention and error rates after deployment, and treating certainty as a decision aid rather than
a guarantee of correctness. Efficiency improvements can also lower the marginal cost of large
scale generation, so deployments should adopt appropriate usage policies when operating at
high volume.

\nocite{langley00}

\bibliography{example_paper}
\bibliographystyle{icml2026}

\newpage
\appendix
\onecolumn
\section{Detailed Model Performance Comparisons}

To evaluate the robustness of Certainty-Guided Reasoning (CGR), we conducted extensive multi-seed evaluations across 64 random seeds for each model architecture: DeepSeek-14B, DeepSeek-70B, and Phi-4. This approach allows us to quantify the stability of the certainty-guided exit mechanism compared to a fixed-budget baseline. As noted in the main text, CGR functions as a precision filter that prioritizes reliability over mere completion.

\subsection{DeepSeek-14B Statistical Stability}

\begin{figure}[H]
    \centering
    \includegraphics[width=0.75\linewidth,trim={0 0 0 22pt},  clip]{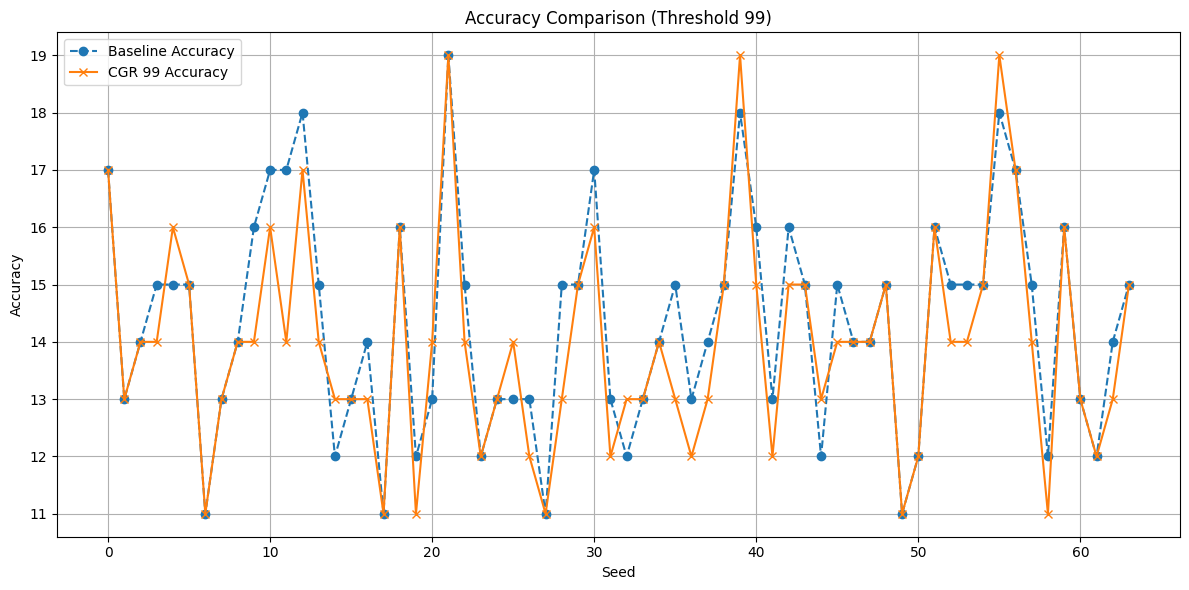}
    \caption{DeepSeek 14b Accuracy Comparison across 64 random seeds.}
    \label{fig:deep14comp}
\end{figure}

\begin{figure}[H]
    \centering
    \includegraphics[width=0.75\linewidth,trim={0 0 0 22pt},  clip]{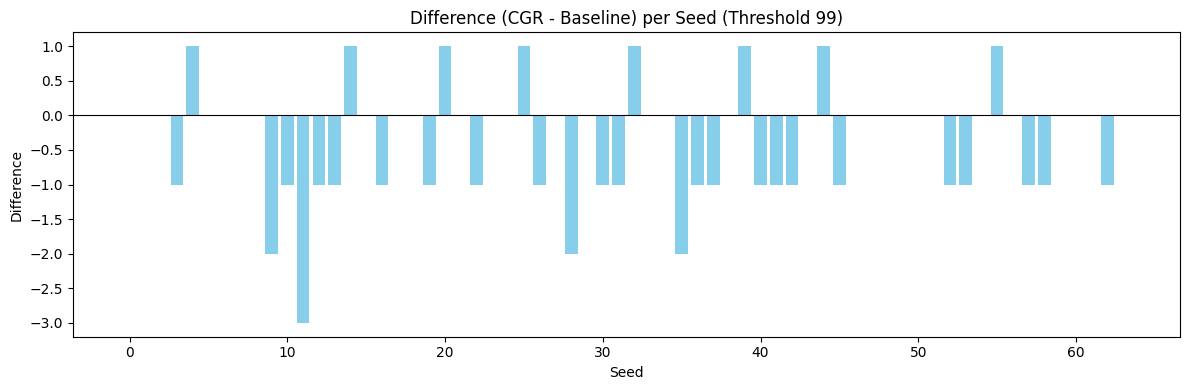}
    \caption{DeepSeek 14b Distance from Baseline (Threshold 0.99) per seed.}
    \label{fig:deep14comp2}
\end{figure}

As shown in Figure \ref{fig:deep14comp}, the accuracy curves for CGR nearly overlap the baseline across the 64-seed evaluation. This indicates that the model's fundamental reasoning capability is preserved even when inference is terminated early due to high certainty. The relative difference plot in Figure \ref{fig:deep14comp2} further demonstrates that most deviations are minor and typically negative, corresponding to uncertain cases where CGR opted to abstain rather than risk a confident error.

\begin{table}[H]
\centering
\caption{CGR Performance on DeepSeek-14B across certainty thresholds (64 seeds).}
\label{tab:deep14_perf}
\begin{tabular}{l ccc cccc}
\hline
\textbf{Threshold} & \textbf{Acc.} & \textbf{Rel. Acc} & \textbf{Diff.} & \textbf{Rel. Diff.} & \textbf{Equal} & \textbf{Better} & \textbf{Worse} \\
\hline
0.96 & 13.41 & 44.69\% & -0.92 & -3.02\% & 18 & 7 & 39 \\
0.97 & 13.58 & 45.26\% & -0.75 & -2.45\% & 20 & 9 & 35 \\
0.98 & 13.77 & 45.89\% & -0.56 & -1.82\% & 23 & 9 & 32 \\
0.99 & 13.98 & 46.61\% & -0.34 & -1.09\% & 31 & 8 & 25 \\
\hline
\end{tabular}
\end{table}

\subsection{Phi-4 Calibration and Invariance}

For Phi-4, accuracy remained effectively constant across thresholds, differing from the baseline by less than 1\%. This stability could be attributed to the model's stronger intrinsic calibration; certainty scores for this model tend to cluster near 0 or 1, leaving the intermediate regions sparsely populated.

\begin{table}[H]
\centering
\caption{CGR Performance on Phi-4 across certainty thresholds (64 seeds).}
\label{tab:phi4_perf}
\begin{tabular}{l ccc cccc}
\hline
\textbf{Threshold} & \textbf{Acc.} & \textbf{Rel. Acc.} & \textbf{Diff.} & \textbf{Rel. Diff.} & \textbf{Equal} & \textbf{Better} & \textbf{Worse} \\
\hline
0.96 & 21.89 & 72.97\% & -0.29 & -0.94\% & 45 & 0 & 18 \\
0.97 & 21.89 & 72.97\% & -0.29 & -0.94\% & 45 & 0 & 18 \\
0.98 & 21.89 & 72.97\% & -0.29 & -0.94\% & 45 & 0 & 18 \\
0.99 & 21.89 & 72.97\% & -0.29 & -0.94\% & 45 & 0 & 18 \\
\hline
\end{tabular}
\end{table}

\begin{figure}[H]
    \centering
    \includegraphics[width=0.75\linewidth,trim={0 0 0 22pt},  clip]{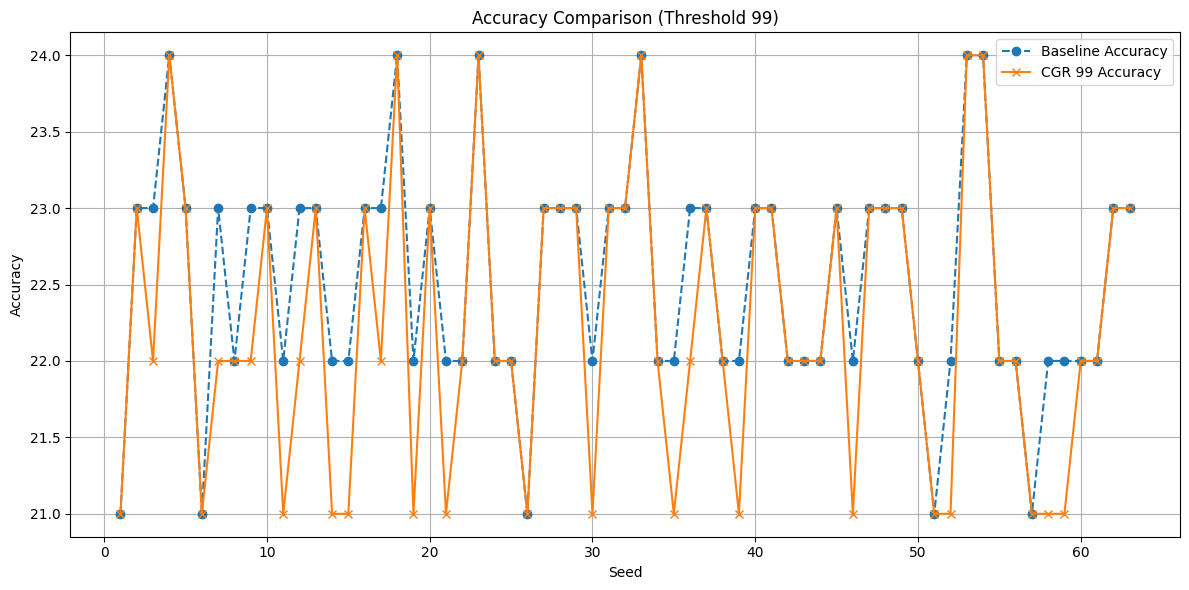}
    \caption{Phi 4 Accuracy Comparison across 64 random seeds.}
    \label{fig:pihcomp}
\end{figure}

\begin{figure}[H]
    \centering
    \includegraphics[width=0.75\linewidth,trim={0 0 0 22pt},  clip]{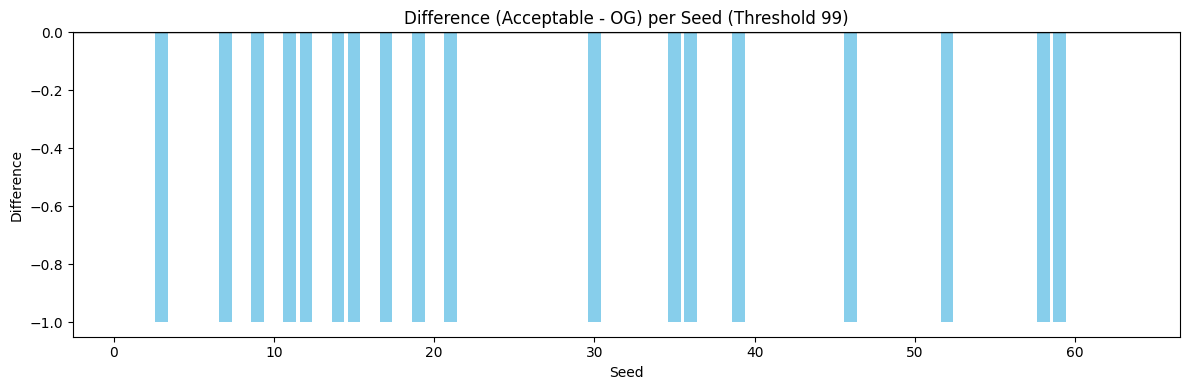}
    \caption{Phi 4 Distance from Baseline (Threshold 0.99) per seed.}
    \label{fig:pihcomp2}
\end{figure}

\subsection{DeepSeek-70B Scalability}

\begin{figure}[H]
    \centering
    \includegraphics[width=0.75\linewidth,trim={0 0 0 22pt},  clip]{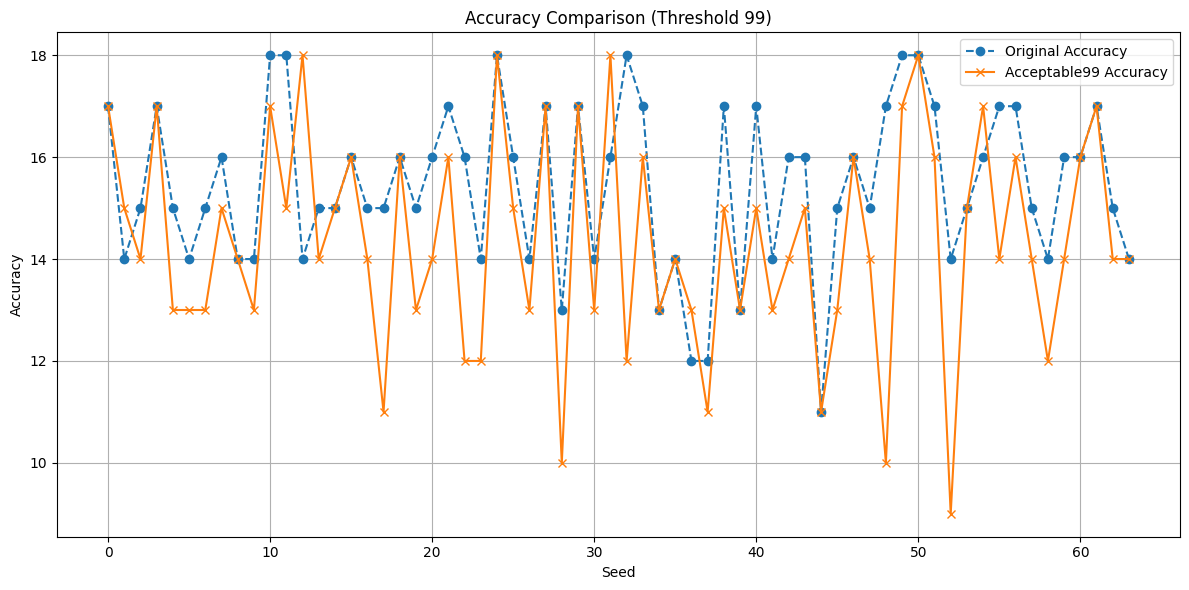}
    \caption{DeepSeek 70B Accuracy Comparison across 64 random seeds.}
    \label{fig:ds70comp}
\end{figure}

\begin{figure}[H]
    \centering
    \includegraphics[width=0.75\linewidth,trim={0 0 0 22pt},  clip]{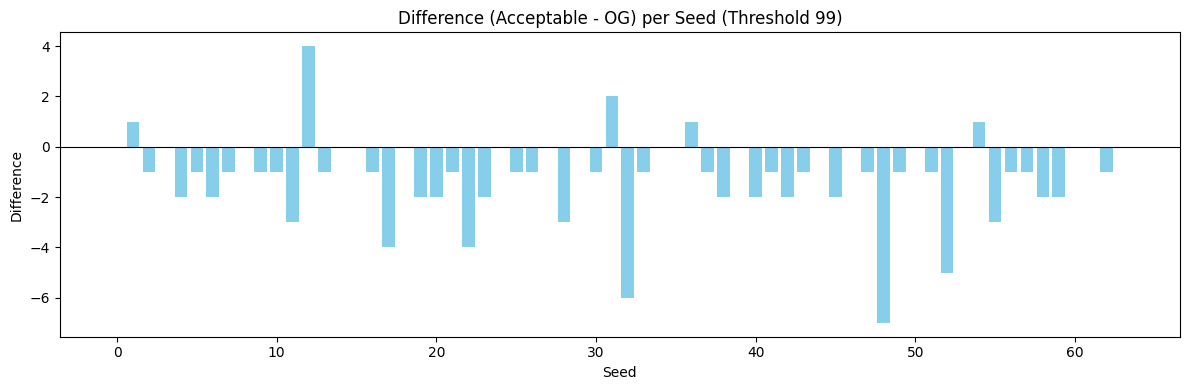}
    \caption{DeepSeek 70B Distance from Baseline (Threshold 0.99) per seed.}
    \label{fig:ds70comp2}
\end{figure}

The results for the largest evaluated model, DeepSeek-70B, closely mirror those of the smaller architectures. Figure \ref{fig:ds70comp} shows that accuracy remains largely identical to the baseline. The minimal divergence shown in Figure \ref{fig:ds70comp2} demonstrates that the CGR mechanism remains meaningful even at large scales.

\newpage
\onecolumn
\section{Grade Metric Sensitivity Analysis}

To evaluate the practical utility of Certainty-Guided Reasoning (CGR) in risk-sensitive environments, we analyze the Grade metric across multiple penalty regimes: $p \in \{0, 0.25, 0.5, 1.0\}$. This analysis reveals the "optimal reasoning depth" for each model, where the balance between accuracy and the risk of incorrect answers is maximized.

\subsection{DeepSeek-14B Grade Performance}

For DeepSeek-14B, the highest average grades consistently occurred between 11k and 15k tokens, as seen in Table \ref{tab:deep14grade}. This suggests that the model often reaches its most reliable conclusion well before the 32k budget limit. As penalties increase, the advantage of CGR-based abstention becomes more pronounced, shifting the optimal stopping point earlier to avoid late-stage reasoning drift.

\begin{figure}[H]
    \centering
    \begin{minipage}{0.48\textwidth}
        \centering
        \includegraphics[width=\linewidth,trim={0 0 0 29pt},  clip]{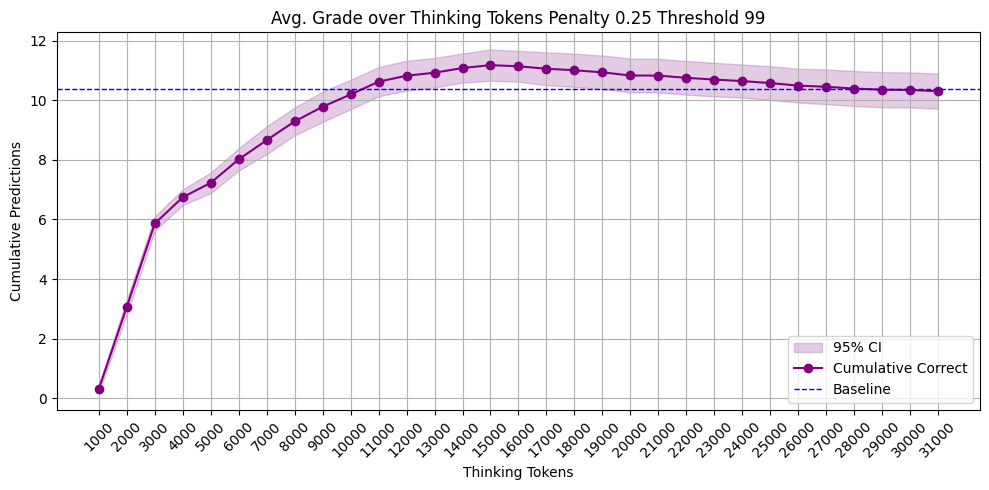}
        \caption{DeepSeek 14b Grade ($p = 0.25$)}
    \end{minipage}\hfill
    \begin{minipage}{0.48\textwidth}
        \centering
        \includegraphics[width=\linewidth,trim={0 0 0 29pt},  clip]{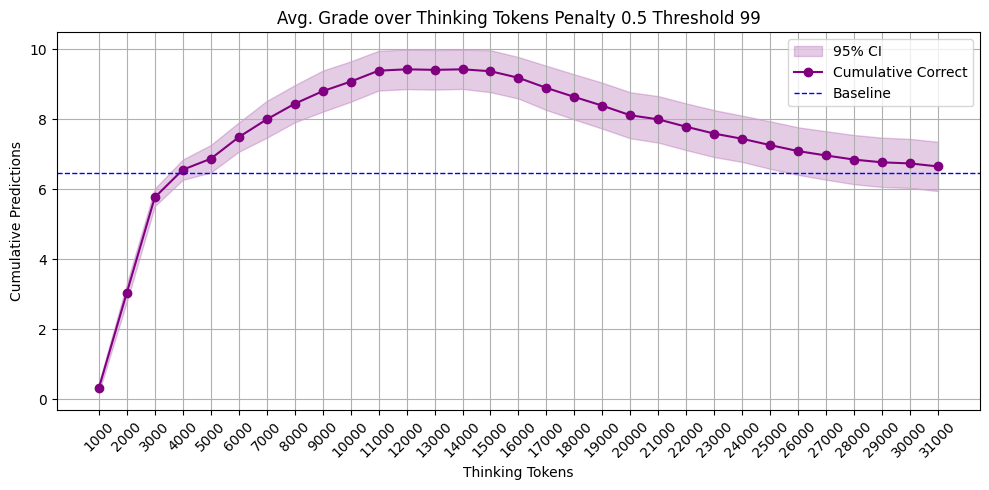}
        \caption{DeepSeek 14b Grade ($p = 0.5$)}
    \end{minipage}
    \label{fig:deep14grade_plots}
\end{figure}

\begin{table}[H]
\centering
\caption{Final Grade values at 31k tokens and optimal stopping points for DeepSeek-14B with CGR ($\theta = 0.99$).}
\label{tab:deep14grade}
\begin{tabular}{lcccccc}
\hline
\textbf{Penalty ($p$)} & \textbf{Final Grade} & \textbf{Baseline} & \textbf{Diff.} & \textbf{Rel. Diff.} & \textbf{Max Grade} & \textbf{Step} \\
\hline
0.00 & 13.98 & 14.31 & -0.34 & -1.15\% & 13.98 & 31000 \\
0.25 & 10.28 & 10.39 & -0.11 & -0.36\% & 11.18 & 15000 \\
0.50 & 6.59 & 6.47 & 0.12 & 0.42\% & 9.42 & 12000 \\
1.00 & -0.78 & -1.38 & 0.59 & 1.98\% & 6.91 & 11000 \\
\hline
\end{tabular}
\end{table}

\subsection{Phi-4 Grade Performance}

Phi-4 exhibits a strong alignment between certainty and correctness, leading to significant Grade improvements in high-penalty settings ($+9.95\%$ for $p=1.0$). Unlike smaller models, Phi-4 reaches stable certainty between 20k and 26k tokens.

\begin{figure}[H]
    \centering
    \begin{minipage}{0.48\textwidth}
        \centering
        \includegraphics[width=\linewidth,trim={0 0 0 29pt},  clip]{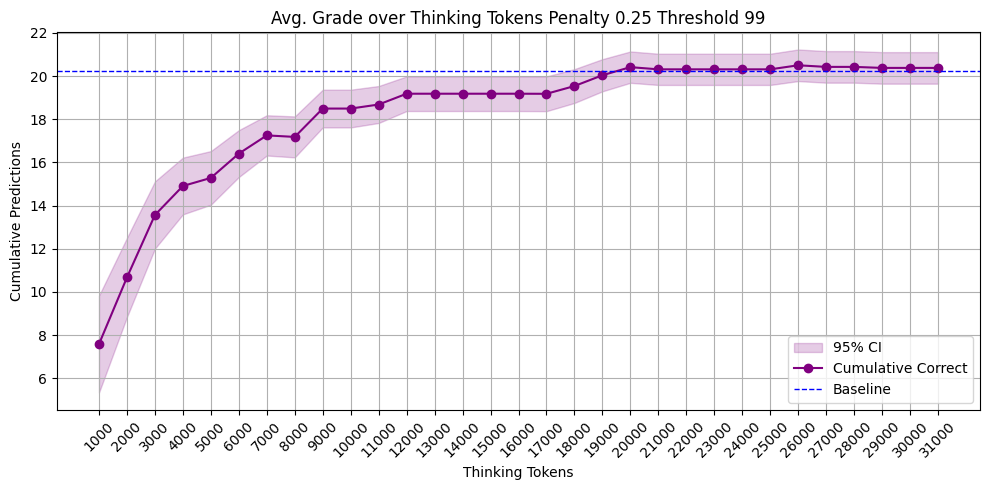}
        \caption{Phi4 Grade ($p = 0.25$)}
    \end{minipage}\hfill
    \begin{minipage}{0.48\textwidth}
        \centering
        \includegraphics[width=\linewidth,trim={0 0 0 29pt},  clip]{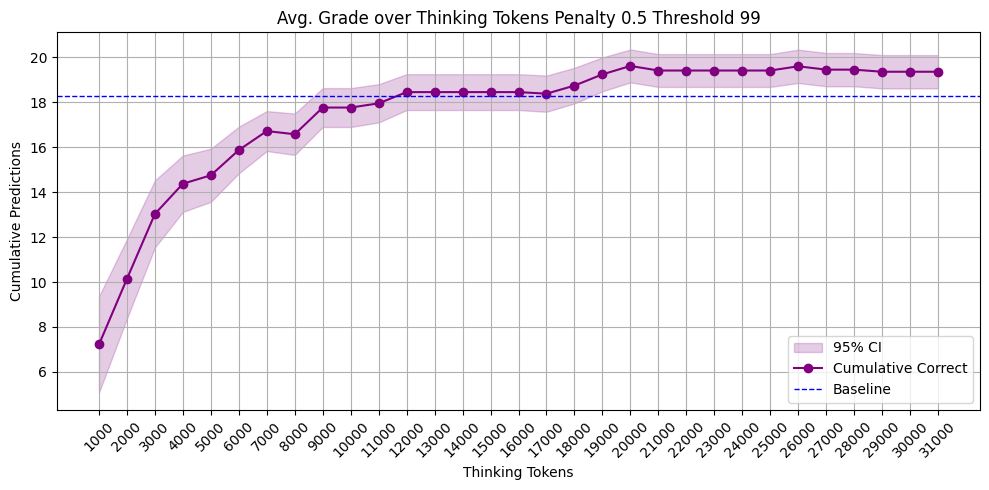}
        \caption{Phi4 Grade ($p = 0.5$)}
    \end{minipage}
    \label{fig:phi4grade_plots}
\end{figure}

\begin{table}[H]
\centering
\caption{Final Grade values at 31k tokens and optimal stopping points for Phi-4 with CGR ($\theta = 0.99$).}
\label{tab:phi4grade}
\begin{tabular}{lcccccc}
\hline
\textbf{Penalty ($p$)} & \textbf{Final Grade} & \textbf{Baseline} & \textbf{Diff.} & \textbf{Rel. Diff.} & \textbf{Max Grade} & \textbf{Step} \\
\hline
0.00 & 21.39 & 22.17 & -0.78 & -2.60\% & 21.39 & 26000 \\
0.25 & 20.38 & 20.21 & 0.16 & 0.53\% & 20.50 & 26000 \\
0.50 & 19.36 & 18.26 & 1.10 & 3.67\% & 19.62 & 20000 \\
1.00 & 17.33 & 14.34 & 2.98 & 9.95\% & 18.03 & 20000 \\
\hline
\end{tabular}
\end{table}




\newpage
\section{Token Savings and Computational Efficiency}

The primary advantage of Certainty-Guided Reasoning (CGR) is its ability to reduce computational waste by terminating reasoning trajectories once a stable, high-confidence answer is reached. This section provides a detailed breakdown of the token savings achieved across different certainty thresholds and model architectures.

\subsection{Threshold-Efficiency Trade-off}

As shown in Table \ref{tab:token_savings}, there is a direct trade-off between the certainty threshold $\theta$ and the resulting token savings. Lower thresholds (e.g., $\theta = 0.96$) allow for more aggressive early exiting, yielding a reduction of over 3.3 million tokens in aggregate across 64 seeds. Stricter thresholds, such as our recommended $\theta = 0.99$, still deliver substantial efficiency gains (over 2.0 million tokens saved) while providing the strongest guarantees of correctness.

\begin{table}[H]
\centering
\caption{Aggregate token savings on AIME2025 across certainty thresholds (64 seeds, DeepSeek-14B).}
\label{tab:token_savings}
\begin{tabular}{cccc}
\hline
\textbf{Threshold} & \textbf{Total Tokens Saved} & \textbf{Avg. per Seed} & \textbf{Avg. per Question} \\
\hline
0.96 & 3,380,578 & 52,821 & 1,760 \\
0.97 & 3,081,690 & 48,151 & 1,605 \\
0.98 & 2,739,761 & 42,808 & 1,426 \\
0.99 & 2,042,389 & 31,912 & 1,063 \\
\hline
\end{tabular}
\end{table}

\subsection{Per-Seed and Per-Question Savings Distribution}

\begin{figure}[H]
    \centering
    \includegraphics[width=0.75\linewidth,trim={0 0 0 22pt},  clip]{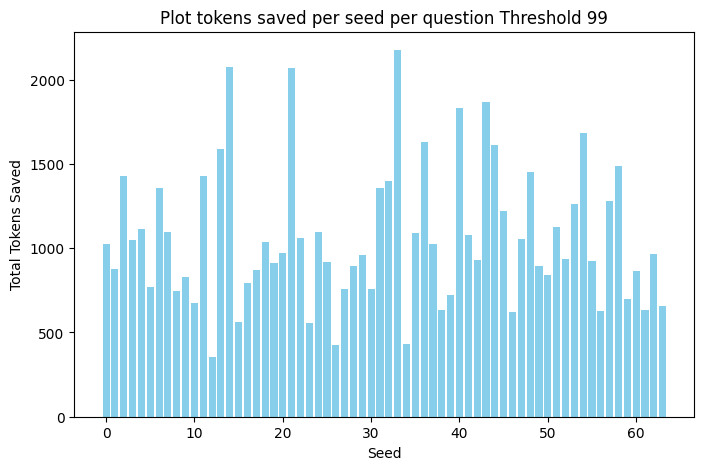}
    \caption{Total tokens saved per seed per question for DeepSeek-14B ($\theta = 0.99$).}
    \label{fig:savings_dist}
\end{figure}

\begin{figure}[H]
    \centering
    \includegraphics[width=0.75\linewidth,trim={0 0 0 22pt},  clip]{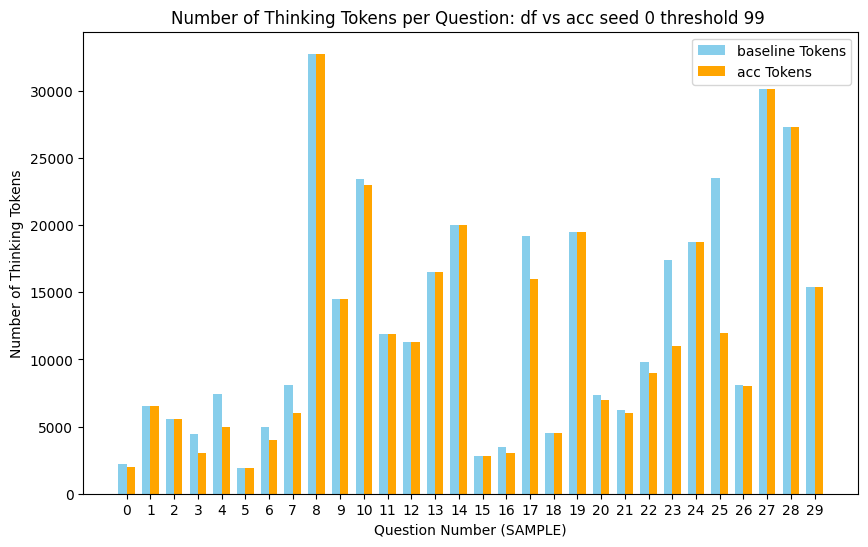}
    \caption{Comparison of total tokens used: Baseline (Fixed 32k) vs. CGR (Adaptive).}
    \label{fig:baseline_vs_cgr}
\end{figure}

Figure \ref{fig:savings_dist} illustrates that token savings are consistently achieved across all 64 random seeds. Figure \ref{fig:baseline_vs_cgr} compares the fixed 32k-token baseline against the adaptive CGR usage, highlighting how the algorithm identifies and exploits "easy" instances that require significantly less than the allocated budget.

\subsection{Token Savings as a Difficulty Proxy}

\begin{figure}[H]
    \centering
    \includegraphics[width=0.75\linewidth,trim={0 0 0 22pt},  clip]{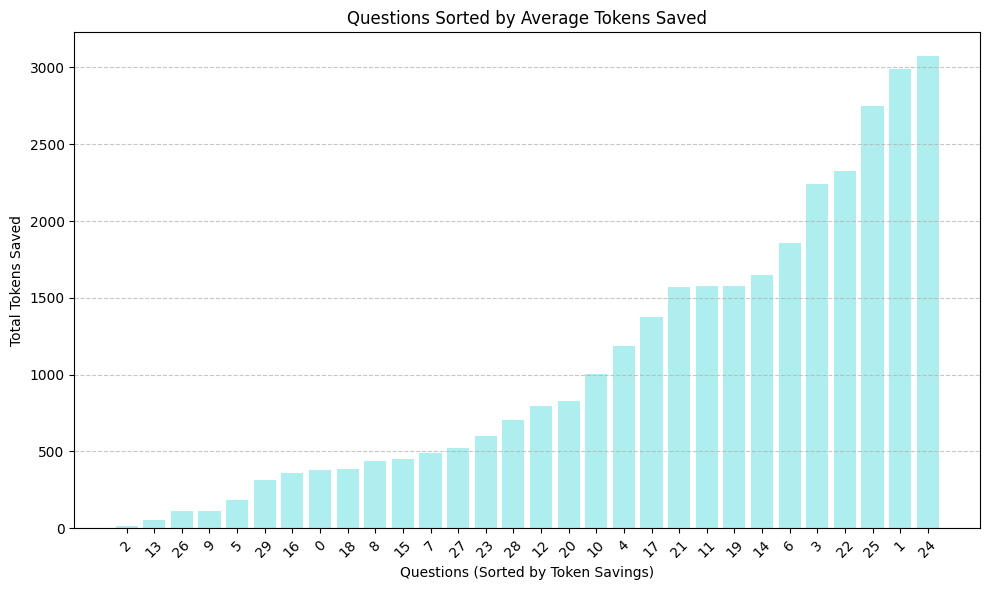}
    \caption{Token savings per question on AIME2025, serving as an indicator of problem difficulty.}
    \label{fig:difficulty_ranking}
\end{figure}

The distribution of token savings also provides a window into problem difficulty for reasoning models. Questions where CGR saves a high percentage of tokens are those where the model converges on a certain answer rapidly, whereas questions with minimal savings represent harder instances requiring full-depth deliberation. This relationship is visualized in Figure \ref{fig:difficulty_ranking}.

\newpage
\section{Experimental Implementation Details}

This appendix provides the necessary technical details to ensure the reproducibility of our results. We describe the core algorithms, the custom logits processing for real-time inference, and the post-hoc simulation protocol used for standardized evaluation.

\subsection{Algorithmic Framework}

The Certainty-Guided Reasoning (CGR) procedure and its variants are formalized in the algorithms below. Algorithm \ref{alg:cgr_algo} describes the standard early-exit procedure. Algorithm \ref{alg:bf_algo} details the Budget Forcing mechanism used to prevent premature model halting. Finally, Algorithm \ref{alg:cgr_bf} presents the unified approach that balances forced exploration with certainty-guided efficiency.

\begin{algorithm}[tbh]
   \caption{Certainty-Guided Reasoning (CGR)}
   \label{alg:cgr_algo}
\begin{algorithmic}
   \STATE {\bfseries Input:} query $q$, model $\mathcal{M}$, budget $B$, threshold $\theta$, interval $\Delta$
   \STATE Initialize reasoning trace $o \leftarrow \emptyset$
   \STATE Initialize token count $t \leftarrow 0$
   \WHILE{$t < B$}
   \STATE Sample next token $x \sim \mathcal{M}(\cdot | q, o)$
   \STATE Append token: $o \leftarrow o \mathbin{\|} x$
   \STATE $t \leftarrow t + 1$
   \IF{$x = \text{\texttt{</think>}}$}
   \STATE {\bfseries break} \COMMENT{Natural model termination}
   \ENDIF
   \IF{$t \pmod{\Delta} = 0$}
   \STATE $c \leftarrow \textsc{CertaintyProbe}(\mathcal{M}, q, o)$ \COMMENT{Periodic probing}
   \IF{$c \geq \theta$}
   \STATE {\bfseries break} \COMMENT{Certainty-guided early exit}
   \ENDIF
   \ENDIF
   \ENDWHILE
   \STATE {\bfseries Output:} $\textsc{DecodeAnswer}(\mathcal{M}, q, o)$
\end{algorithmic}
\end{algorithm}

\begin{algorithm}[tb]
   \caption{Budget Forcing Implementation}
   \label{alg:bf_algo}
\begin{algorithmic}
   \STATE {\bfseries Input:} query $q$, model $\mathcal{M}$, budget $B$
   \STATE Initialize reasoning trace $o \leftarrow \emptyset$
   \STATE Initialize token count $t \leftarrow 0$
   \WHILE{$t < B$}
   \STATE Sample next token $x \sim \mathcal{M}(\cdot | q, o)$
   \IF{$x = \text{\texttt{</think>}}$}
   \STATE $x \leftarrow \text{``\texttt{\textbackslash nWait}''}$ \COMMENT{Override natural stop token}
   \ENDIF
   \STATE Append token: $o \leftarrow o \mathbin{\|} x$
   \STATE $t \leftarrow t + 1$
   \ENDWHILE
   \STATE Append ``Final Answer: \verb|\boxed{|'' to $o$
   \STATE {\bfseries Output:} $\textsc{ExtractAnswer}(o)$
\end{algorithmic}
\end{algorithm}

\begin{algorithm}[tb]
   \caption{CGR with Budget Forcing}
   \label{alg:cgr_bf}
\begin{algorithmic}
   \STATE {\bfseries Input:} query $q$, model $\mathcal{M}$, budget $B$, threshold $\theta$, interval $\Delta$
   \STATE Initialize reasoning trace $o \leftarrow \emptyset$
   \STATE Initialize token count $t \leftarrow 0$
   \WHILE{$t < B$}
   \STATE Sample next token $x \sim \mathcal{M}(\cdot | q, o)$
   \IF{$x = \text{\texttt{</think>}}$}
   \STATE $c \leftarrow \textsc{CertaintyProbe}(\mathcal{M}, q, o)$
   \IF{$c \geq \theta$}
   \STATE {\bfseries break} \COMMENT{Allow stop if certainty is high}
   \ELSE
   \STATE $x \leftarrow \text{``\texttt{\textbackslash nWait}''}$ \COMMENT{Force continue if uncertain}
   \ENDIF
   \ENDIF
   \STATE Append token: $o \leftarrow o \mathbin{\|} x$
   \STATE $t \leftarrow t + 1$
   \IF{$t \pmod{\Delta} = 0$}
   \STATE $c \leftarrow \textsc{CertaintyProbe}(\mathcal{M}, q, o)$
   \IF{$c \geq \theta$}
   \STATE {\bfseries break}
   \ENDIF
   \ENDIF
   \ENDWHILE
   \STATE {\bfseries Output:} $\textsc{DecodeAnswer}(\mathcal{M}, q, o)$
\end{algorithmic}
\end{algorithm}

\subsection{Implementation Specs}

\textbf{Certainty Threshold and Interval:} We selected $\theta = 0.99$ as the default operational threshold based on empirical validation showing it yields the optimal balance between accuracy and efficiency. Probing is conducted every 1,000 tokens to minimize overhead while maintaining responsiveness to reasoning progress.

\textbf{Custom Logits Processor:} For real-time inference control, we implemented a custom logits processor. This processor monitors the probability distribution at each step and programmatically suppresses the \texttt{</think>} token by setting its logit value to $-\infty$ (\texttt{-torch.inf}) whenever the budget forcing conditions are active.

\textbf{Reproducibility and Post-Hoc Analysis:} Due to non-determinism in long-form generation caused by floating-point rounding errors, we adopted a post-hoc analysis method for comparative benchmarking. We first generated a full 32,000-token trace for each instance and subsequently "cut" the trace at 1,000-token intervals to simulate early exits. This ensures that any performance variation between thresholds is solely attributable to the stopping rule and not to stochastic variations in the reasoning trace.

\end{document}